\documentclass[sigconf]{acmart}
\usepackage{color}
\usepackage{multicol}
\usepackage{url}
\usepackage{bbm}
\usepackage{verbatim} 
\usepackage{subfigure} 
\usepackage{multirow}
\usepackage{algorithm}
\usepackage[noend]{algpseudocode}
\usepackage{xspace}
\usepackage{graphicx}
\usepackage{epsfig}
\usepackage{amsmath}
\usepackage{amssymb}
\usepackage{enumerate}
\usepackage{enumitem}

\usepackage{dcolumn}
\usepackage{hhline}

\usepackage{graphicx,psfrag,amsmath,amsfonts,appendix}
\usepackage{algorithm}

\newcommand{\reals}{{\mbox{\bf R}}}

\newcommand{\hide}[1]{}
\newcommand{\xhdr}[1]{\vspace{1.7mm}\noindent{{\bf #1.}}}

\newcommand{\ie}{\emph{i.e.}}

\newcommand{\norm}[1]{\left\|#1\right\|}

\newcommand{\fronorm}[1]{\norm{#1}_{F}}

\newcommand{\Tr}[1]{\mathop{\mathbf{Tr}(#1)}}

\newcommand{\argmin}[1]{\underset{#1}{\mathop{\rm argmin ~}}}

\graphicspath{{./FIG/}}

\newfont{\mycrnotice}{ptmr8t at 7pt}
\newfont{\myconfname}{ptmri8t at 7pt}

\begin{document}

\copyrightyear{2017} 
\acmYear{2017} 
\setcopyright{acmlicensed}
\acmConference{KDD '17}{August 13-17, 2017}{Halifax, NS, Canada}\acmPrice{15.00}\acmDOI{10.1145/3097983.3098060}
\acmISBN{978-1-4503-4887-4/17/08}

\fancyhead{}
\settopmatter{printacmref=false, printfolios=false}

\title{Toeplitz Inverse Covariance-Based Clustering of\\ Multivariate Time Series Data}

\author{David Hallac, Sagar Vare, Stephen Boyd, Jure Leskovec}
\affiliation{%
  \institution{Stanford University}
}
\email{{hallac, svare, boyd, jure}@stanford.edu}

\begin{abstract}

Subsequence clustering of multivariate time series is a useful tool for discovering repeated patterns in temporal data. Once these patterns have been discovered, seemingly complicated datasets can be interpreted as a temporal sequence of only a small number of states, or \emph{clusters}. For example, raw sensor data from a fitness-tracking application can be expressed as a timeline of a select few actions (\ie, walking, sitting, running). However, discovering these patterns is challenging because it requires simultaneous segmentation and clustering of the time series. Furthermore, interpreting the resulting clusters is difficult, especially when the data is high-dimensional. Here we propose a new method of model-based clustering, which we call \emph{Toeplitz Inverse Covariance-based Clustering} (TICC). Each cluster in the TICC method is defined by a correlation network, or Markov random field (MRF), characterizing the interdependencies between different observations in a typical subsequence of that cluster. Based on this graphical representation, TICC simultaneously segments and clusters the time series data. We solve the TICC problem through alternating minimization, using a variation of the expectation maximization (EM) algorithm. We derive closed-form solutions to efficiently solve the two resulting subproblems in a scalable way, through dynamic programming and the alternating direction method of multipliers (ADMM), respectively. We validate our approach by comparing TICC to several state-of-the-art baselines in a series of synthetic experiments, and we then demonstrate on an automobile sensor dataset how TICC can be used to learn interpretable clusters in real-world scenarios.

\end{abstract}

\maketitle




\section{Introduction}
\label{sec:intro}

Many applications, ranging from automobiles~\cite{miyajima2007driver} to financial markets~\cite{N:11} and wearable sensors~\cite{morchen2005extracting}, generate large amounts of time series data. In most cases, this data is multivariate, where each timestamped observation consists of readings from multiple entities, or \emph{sensors}. 
These long time series can often be broken down into a sequence of states, each defined by a simple ``pattern'', where the states can reoccur many times. For example, raw sensor data from a fitness-tracking device can be interpreted as a temporal sequence of actions~\cite{parkka2006activity} (\ie, walking for 10 minutes, running for 30 minutes, sitting for 1 hour, then running again for 45 minutes). Similarly, using automobile sensor data, a single driving session can be expressed as a sequential timeline of a few key states: turning, speeding up, slowing down, going straight, stopping at a red light, etc. 
This representation can be used to discover repeated patterns, understand trends, detect anomalies and more generally, better interpret large and high-dimensional datasets.

\begin{figure}[t]
  \centering
  \includegraphics[width=\linewidth]{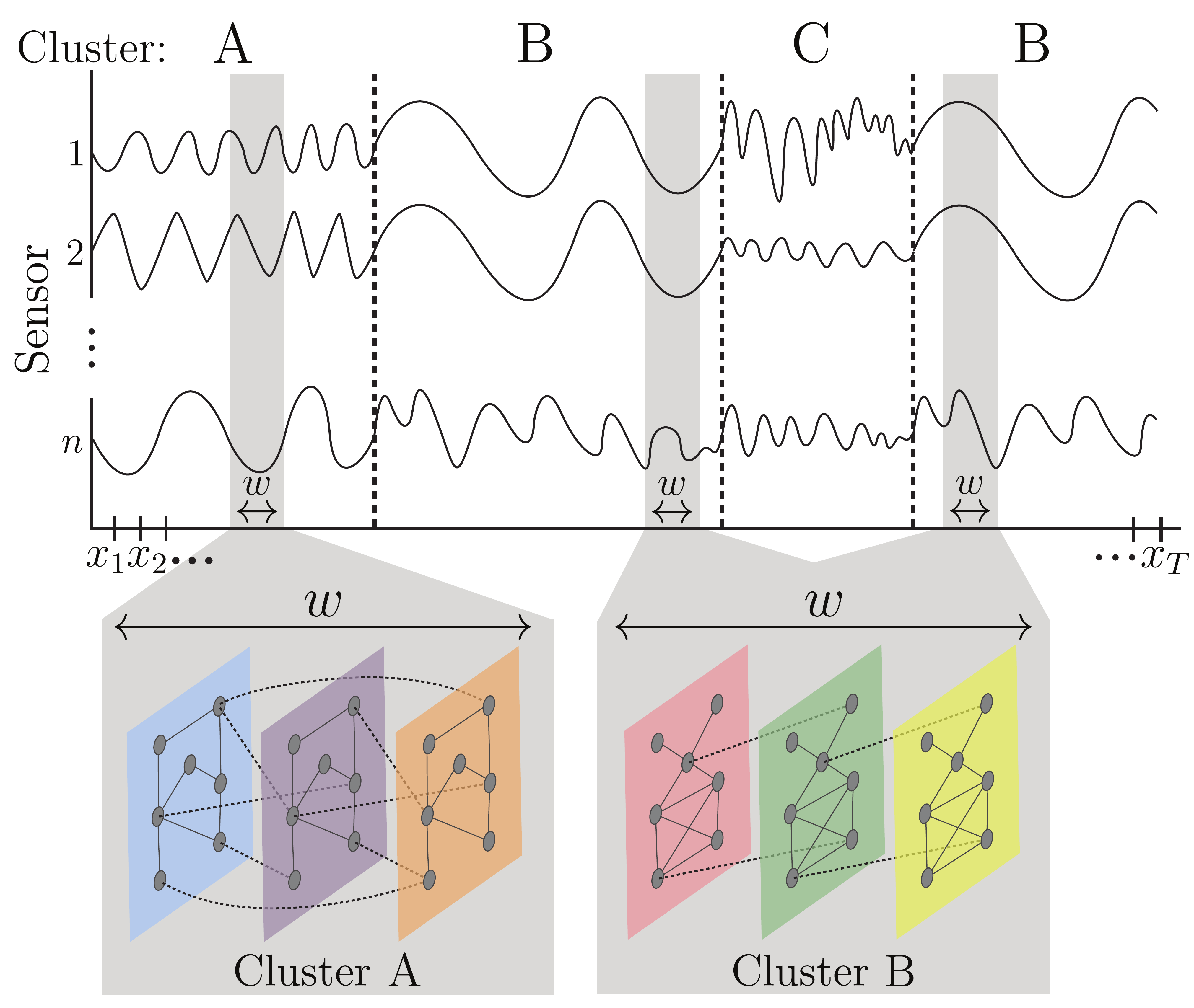}
  \vspace{-5mm}
  \caption{Our TICC method segments a time series into a sequence of states, or ``clusters'' (\ie, A, B, or C). Each cluster is characterized by a correlation network, or MRF, defined over a short window of size $w$. This MRF governs the (time-invariant) partial correlation structure of \emph{any} window inside a segment belonging to that cluster. Here, TICC learns both the cluster MRFs and the time series segmentation.}
  \vspace{-4mm}
  \label{fig:musical}
  \label{fig:multimodal}
\end{figure}

To achieve this representation, it is necessary to simultaneously segment and cluster the time series.
This problem is more difficult than standard time series segmentation~\cite{hallac2016greedy,himberg2001time}, since multiple segments can belong to the same cluster. However, it is also harder than subsequence clustering~\cite{begum2015accelerating, smyth1997clustering} 
because each data point cannot be clustered independently 
(since neighboring points are encouraged to belong to the same cluster).
Additionally, even if one is able to simultaneously segment and cluster the data, the question still arises as to how to interpret the different clusters. These clusters are rarely known a priori, and thus are best learned through data. However, without prior knowledge, it is difficult to understand what each of the clusters refers to. 
Traditional clustering methods are not particularly well-suited to discover interpretable structure in the data. This is because they typically rely on distance-based metrics, such as dynamic time warping~\cite{berndt1994using}. These methods focus on matching the raw values, rather than looking for more nuanced structural similarities in the data, for example how different sensors in a car correlate with each other across time.

In this paper, we propose a new method for multivariate time series clustering, which we call \emph{Toeplitz inverse covariance-based clustering} (TICC). In our method, we define each cluster as a dependency network showing the relationships between the different sensors in a short (time-invariant) subsequence (Figure \ref{fig:musical}). For example, in a cluster corresponding to a ``turn'' in an automobile, this network, known as a Markov random field (MRF), might show how the brake pedal at a generic time $t$ might affect the steering wheel angle at time $t+1$. Here, the MRF of a different cluster, such as ``slowing down'', will have a very different dependency structure between these two sensors.
In these MRFs, an edge represents a partial correlation 
between two variables~\cite{KF:09, RH:05, WK:13}. It is important to note that MRFs denote a relationship much stronger than a simple correlation; partial correlations are used to control for the effect of other confounding variables, so the existence of an edge in an MRF implies that there is a \emph{direct} dependency between two variables in the data. Therefore, an MRF provides interpretable insights as to precisely what the key factors and relationships are that characterize each cluster.

In our TICC method, we learn each cluster's MRF by estimating a sparse Gaussian inverse covariance matrix~\cite{FHT:08,YL:06}. With an inverse covariance $\Theta$, if $\Theta_{i,j} = 0$, then by definition, elements $i$ and $j$ in $\Theta$ are conditionally independent (given the values of all other variables).
Therefore, $\Theta$ defines the adjacency matrix of the MRF dependency network~\cite{BED:08,WJ:06}. 
This network has multiple layers, with edges both within a layer and across different layers. 
Here, the number of layers corresponds to the window size of a short subsequence that we define our MRF over.
For example, the MRFs corresponding to clusters A and B in Figure \ref{fig:multimodal} both have three layers.
This multilayer network represents the time-invariant correlation structure of \emph{any} window of observations inside a segment belonging to that cluster.
We learn this structure for each cluster by solving a constrained inverse covariance estimation problem, which we call the \emph{Toeplitz graphical lasso}. The constraint we impose ensures that the resulting multilayer network has a block Toeplitz structure~\cite{gray2006toeplitz} (\ie, any edge between layers $l$ and $l+1$ also exists between layers $l+1$ and $l+2$).
This Toeplitz constraint ensures that our cluster definitions are time-invariant, so the clustering assignment does not depend on the exact starting position of the subsequence. Instead, we cluster this short subsequence based solely on the structural state that the time series is currently in. 

To solve the TICC problem, we use an expectation maximization (EM)-like approach, based on alternating minimization, where we iteratively cluster the data and then update the cluster parameters. Even though TICC involves solving a highly non-convex maximum likelihood problem, our method is able to find a (locally) optimal solution very efficiently in practice. 
When assigning the data to clusters, we have an additional goal of \emph{temporal consistency}, the idea that adjacent data in the time series is encouraged to belong to the same cluster. However, this yields a combinatorial optimization problem.  
We develop a scalable solution using dynamic programming, 
which allows us to efficiently learn the optimal assignments (it takes just $O(KT)$ time to assign the $T$ points into $K$ clusters).
Then, to solve for the cluster parameters, we develop an algorithm to solve the Toeplitz graphical lasso problem. Since learning this graphical structure from data is a computationally expensive semidefinite programming problem~\cite{BV:04,hsieh2013big}, we develop a specialized message-passing algorithm based on the alternating direction method of multipliers (ADMM)~\cite{BPCPE:11}.
In our Toeplitz graphical lasso algorithm, we derive closed-form updates for each of the ADMM subproblems to significantly speed up the solution time.

We then implement our TICC method and apply it to both real and synthetic datasets. We start by evaluating performance on several synthetic examples, where there are known ground truth clusters. We compare TICC with several state-of-the-art time series clustering methods, outperforming them all by at least $41\%$ in terms of cluster assignment accuracy. We also quantify the amount of data needed for accurate cluster recovery for each method, and we see that TICC requires 3x fewer observations than the next best method to achieve similar performance.
Additionally, we discover that our approach is able to accurately reconstruct the underlying MRF dependency network, with an $F_1$ network recovery score between 0.79 and 0.90 in our experiments.
We then analyze an automobile sensor dataset to see an example of how TICC can be used to learn interpretable insights from real-world data. Applying our method, we discover that the automobile dataset has five true clusters, each corresponding to a ``state'' that cars are frequently in. We then validate our results by examining the latitude/longitude locations of the driving session, along with the resulting clustering assignments, to show how TICC can be a useful tool for unsupervised learning from multivariate time series.

\xhdr{Related Work}
This work relates to recent advancements in time series clustering and convex optimization. Subsequence clustering of time series data is a well-developed field. Methods include several variations of dynamic time warping~\cite{begum2015accelerating, keogh2002exact, keogh2000scaling, rakthanmanon2012searching}, symbolic representations~\cite{lin2003symbolic, lin2007experiencing}, and rule-based motif discovery~\cite{das1998rule, li2012visualizing}. There has also been work on simultaneous clustering and segmentation of time series data, which is known as time point clustering~\cite{gionis2003finding, zolhavarieh2014review}.
However, these methods generally rely on distance-based metrics, which in certain situations have been shown to yield unreliable results~\cite{keogh2003clustering}. Instead, our TICC method is a model-based clustering approach, similar to clustering based on ARMA~\cite{xiong2004time}, Gaussian Mixture~\cite{fraley2006mclust}, or hidden Markov models~\cite{smyth1997clustering}. To the best of our knowledge, our method is the first to perform time series clustering based on the graphical dependency structure of each subsequence. This provides interpretability to our clusters, prevents overfitting, and, as we show in Sections 6 and 7, allows us to discover types of patterns that other approaches are unable to find. We do so by proposing a structured inverse covariance estimation problem, which we call the Toeplitz graphical lasso. This problem is a variation on the well-known graphical lasso problem~\cite{FHT:08} where we enforce a block Toeplitz structure on the solution. While many algorithms exist to solve the standard graphical lasso~\cite{BED:08, hsieh2013big, hsieh2014quic}, we are not aware of any methods specifically adapted for the block Toeplitz case. We propose an ADMM approach because the overall optimization problem can be split into ADMM-friendly subproblems, where we can derive closed-form proximal operators~\cite{BPCPE:11} to quickly solve the optimization problem.

\section{Problem Setup}
\label{sec:original}

Consider a time series of $T$ sequential observations,
\begin{equation*}
\small
x_{\mathrm{orig}} = \begin{bmatrix}
	| & | & | & & | \\
    x_{1} & x_{2} & x_{3} & \dots  & x_{T} \\
	| & | & | & & |
	\end{bmatrix},
\end{equation*}
where $x_i \in \reals^n$ is the $i$-th multivariate observation.  
Our goal is to cluster these $T$ observations into $K$ clusters.
However, instead of clustering each observation in isolation, we treat each point in the context of its predecessors in the time series. Thus, rather than just looking at $x_t$, we instead cluster a short subsequence of size $w \ll T$ that ends at $t$. This consists of observations $x_{t-w+1},\ldots,x_{t}$, which we concatenate into an $nw$-dimensional vector that we call $X_t$. 
We refer to this new sequence, from $X_1$ to $X_T$, as $X$. Note that there is a bijection, or a bidirectional one-to-one mapping, between each point $x_t$ and its resulting subsequence $X_t$. (The first $w$ observations of $x_{\mathrm{orig}}$ simply map to a shorter subsequence, since the time series does not start until $x_1$.)
These subsequences are a useful tool to provide proper context for each of the $T$ observations. For example, in automobiles, a single observation may show the current state of the car (\ie, driving straight at 15mph), but a short window, even one that lasts just a fraction of a second, allows for a more complete understanding of the data (\ie, whether the car is speeding up or slowing down).
As such, rather than clustering the observations directly, our approach instead consists of clustering these subsequences $X_1, \ldots, X_T$. We do so in such a way that encourages adjacent subsequences to belong to the same cluster, a goal that we call \emph{temporal consistency}. Thus, our method can be viewed as a form of time point clustering~\cite{zolhavarieh2014review}, where we simultaneously segment and cluster the time series. 

\xhdr{Toeplitz Inverse Covariance-Based Clustering (TICC)}
We define each cluster by a Gaussian inverse covariance $\Theta_i \in \reals^{nw\times nw}$. Recall that inverse covariances show the conditional independency structure between the variables~\cite{KF:09}, so $\Theta_i$ defines a Markov random field encoding the structural representation of cluster $i$. In addition to providing interpretable results, sparse graphical representations are a useful way to prevent overfitting \cite{L:96}. As such, our objective is to solve for these $K$ inverse covariances $\mathbf{\Theta} = \{ \Theta_1, \ldots, \Theta_K \}$, one per cluster, and the resulting assignment sets $\mathbf{P} = \{ P_1, \ldots, P_K \}$, where $P_i \subset \{1,2,\ldots,T\}$. Here, each of the $T$ points are assigned to exactly one cluster. Our overall optimization problem is
\begin{align}
\small
\argmin{\mathbf{\Theta} \in \mathcal{T}, \mathbf{P}} &\sum_{i=1}^{K} \Bigg[ \overbrace{\|\lambda \circ \Theta_i\|_1}^{\text{\normalfont sparsity}} +
 \sum_{X_t \in P_i} \left( \overbrace{-\ell\ell(X_t,\Theta_i)}^{\text{\normalfont log likelihood}} + \overbrace{\beta \mathbbm{1}\{X_{t-1} \not\in P_i\}}^{\text{\normalfont temporal consistency}} \right)\Bigg].
\label{overallProb}
\end{align}
We call this the \emph{Toeplitz inverse covariance-based clustering} (TICC) problem. Here, $\mathcal{T}$ is the set of symmetric block Toeplitz $nw \times nw$ matrices 
and $\|\lambda \circ \Theta_i\|_1$ is an $\ell_1$-norm penalty of the Hadamard (element-wise) product to incentivize a sparse inverse covariance (where $\lambda \in \reals^{nw \times nw}$ is a regularization parameter). Additionally, $\ell\ell(X_t,\Theta_i)$ is the log likelihood that $X_t$ came from cluster $i$,
\begin{align}
\small
\ell\ell(X_t,\Theta_i) = -\frac{1}{2} &(X_t - \mu_i)^T \Theta_i (X_t - \mu_i) \nonumber \\
 &+ \frac{1}{2} \log\det \Theta_i - \frac{n}{2}\log(2\pi),
 \label{loglikelihood}
\end{align}
where $\mu_i$ is the empirical mean of cluster $i$. In Problem \eqref{overallProb}, $\beta$ is a parameter that enforces temporal consistency, and $\mathbbm{1}\{X_{t-1} \not\in P_i\}$ is an indicator function checking whether neighboring points are assigned to the same cluster.

\xhdr{Toeplitz Matrices}
Note that we constrain the $\Theta_i$'s, the inverse covariances, to be block Toeplitz. Thus, each $nw \times nw$ matrix can be expressed in the following form,
\begin{align*}
\small
\Theta_i = 
\left[
\begin{array}{c c c c c c}
A^{(0)} & (A^{(1)})^T & (A^{(2)})^T & \cdots & \cdots & (A^{(w-1)})^T \\
A^{(1)} & A^{(0)} & (A^{(1)})^T & \ddots & & \vdots \\
A^{(2)} & A^{(1)} & \ddots & \ddots & \ddots & \vdots \\
\vdots & \ddots & \ddots & \ddots & (A^{(1)})^T & (A^{(2)})^T \\
\vdots & & \ddots & A^{(1)} & A^{(0)} & (A^{(1)})^T \\
A^{(w-1)} & \cdots & \cdots & A^{(2)} & A^{(1)} & A^{(0)} \\
\end{array}
\right],
\end{align*}
where $A^{(0)}, A^{(1)}, \ldots, A^{(w-1)} \in \reals^{n\times n}$. 
Here, the $A^{(0)}$ sub-block represents the intra-time partial correlations, so $A^{(0)}_{ij}$ refers to the relationship between concurrent values of sensors $i$ and $j$. In the MRF corresponding to this cluster, $A^{(0)}$ defines the adjacency matrix of the edges within each layer. On the other hand, the off-diagonal sub-blocks refer to ``cross-time'' edges. For example, $A^{(1)}_{ij}$ shows how sensor $i$ at some time $t$ is correlated to sensor $j$ at time $t+1$, and $A^{(2)}$ shows the edge structure between time $t$ and time $t+2$. The block Toeplitz structure of the inverse covariance means that we are making a time-invariance assumption over this length-$w$ window (we typically expect this window size to be much smaller than the average segment length). As a result, in Figure \ref{fig:multimodal}, for example, the edges between layer 1 and layer 2 must also exist between layers 2 and 3.
We use this assumption because we are looking for a unique structural pattern to identify each cluster. We consider each cluster to be a certain ``state''. When the time series is in this state, it retains a certain (time-invariant) structure that persists throughout this segment, regardless of the window's starting point. By enforcing a Toeplitz structure on the inverse covariance, we are able to model this time invariance and incorporate it into our estimate of $\Theta_i$.

\xhdr{Regularization Parameters}
Our TICC optimization problem has two regularization parameters: $\lambda$, which determines the sparsity level in the MRFs characterizing each cluster, and $\beta$, the smoothness penalty that encourages adjacent subsequences to be assigned to the same cluster. Note that even though $\lambda$ is a $nw\times nw$ matrix, we typically set all its values to a single constant, reducing the search space to just one parameter. In applications where there is prior knowledge as to the proper sparsity or temporal consistency, $\lambda$ and $\beta$ can be chosen by hand. More generally, the parameter values can also be selected by a more principled method, such as Bayesian information criterion (BIC)~\cite{HTF:09} or cross-validation.

\xhdr{Window Size}
Recall that instead of clustering each point $x_t$ in isolation, we cluster a short window, or subsequence, going from time $t-w+1$ to $t$, which we concatenate into a $nw$-dimensional vector that we call $X_t$. The Toeplitz constraint assumes that each cluster has a time-invariant structure, but this window size is still a relevant parameter. In particular, it allows us to learn cross-time correlations (\ie, sensor $i$ at time $t$ affects sensor $j$ at time $t+1$). 
The larger the window, the farther these cross-time edges can reach. 
However, we do not want our window to be too large, since it may struggle to properly classify points at the segment boundaries, where our time-invariant assumption may not hold.
For this reason, we generally keep the value of $w$ relatively small. However, its exact value should generally be chosen depending on the application, the granularity of the observations, and the average expected segment length. It can also be selected via BIC or cross-validation, though as we discover in Section 6, our TICC algorithm is relatively robust to the selection of this window size parameter.

\xhdr{Selecting the Number of Clusters}
As with many clustering algorithms, the number of clusters $K$ is an important parameter in TICC. There are various methods for doing so. If there is some labeled ground truth data, we can use cross-validation on a test set or normalized mutual information~\cite{cover2012elements} to evaluate performance. 
If we do not have such data, we can use BIC or the silhouette score~\cite{rousseeuw1987silhouettes} to select this parameter.
However, the exact number of clusters will often depend on the application itself, especially since we are also looking for interpretability in addition to accuracy.

\section{Alternating Minimization}
\label{sec:streaming}

Problem \eqref{overallProb} is a mixed combinatorial and continuous optimization problem. There are two sets of variables, the cluster assignments $P$ and inverse covariances $\Theta$, both of which are coupled together to make the problem highly non-convex. 
As such, there is no tractable way to solve for the globally optimal solution. Instead, we use a variation of the expectation maximization (EM) algorithm to alternate between assigning points to clusters and then updating the cluster parameters. While this approach does not necessarily reach the global optimum, similar types of methods have been shown to perform well on related problems~\cite{fraley2006mclust}. Here, we define the subproblems that comprise the two steps of our method. Then, in Section 4, we derive fast algorithms to solve both subproblems and formally describe our overall algorithm to solve the TICC problem.

\subsection{Assigning Points to Clusters}
We assign points to clusters by fixing the value of $\mathbf{\Theta}$ and solving the following combinatorial optimization problem for $\mathbf{P} = \{ P_1, \ldots, P_K \}$, 
\begin{align}
\small
\mathrm{minimize} \sum_{i=1}^K \sum_{X_t \in P_i} -\ell\ell(X_t,\Theta_i) + \beta \mathbbm{1}\{X_{t-1} \not\in P_i\}. 
\label{clustAssign}
\end{align}
This problem assigns each of the $T$ subsequences to one of the $K$ clusters to jointly maximize the log likelihood and the temporal consistency, with the tradeoff between the two objectives regulated by the regularization parameter $\beta$. When $\beta = 0$, the subsequences $X_1, \ldots, X_T$ can all be assigned independently, since there is no penalty to encourage neighboring subsequences to belong to the same cluster. This can be solved by simply assigning each point to the cluster that maximizes its likelihood. As $\beta$ gets larger, neighboring subsequences are more and more likely to be assigned to the same cluster. As $\beta \rightarrow \infty$, the switching penalty becomes so large that all the points in the time series are grouped together into just one cluster. Even though Problem \eqref{clustAssign} is combinatorial, we will see in Section 4.1 that we can use dynamic programming to efficiently find the globally optimal solution for this TICC subproblem.

\subsection{Toeplitz Graphical Lasso}
Given the point assignments $\mathbf{P}$, our next task is to update the cluster parameters $\Theta_1, \ldots, \Theta_K$ by solving Problem \eqref{overallProb} while holding $\mathbf{P}$ constant. We can solve for each $\Theta_i$ in parallel.
To do so, we notice that we can rewrite the negative log likelihood in Problem \eqref{loglikelihood} in terms of each $\Theta_i$. This likelihood can be expressed as 
\begin{align*}
\small
\sum_{X_t \in P_i} -\ell\ell(X_t,\Theta_i) = -|P_i|(\log\det\Theta_i + \mathrm{tr}(S_i \Theta_i)) + C, 
\end{align*}
where $|P_i|$ is the number of points assigned to cluster $i$, $S_i$ is the empirical covariance of these points, and $C$ is a constant that does not depend on $\Theta_i$. Therefore, the M-step of our EM algorithm is 
\begin{align}
\small
&\mathrm{minimize}\qquad -\log\det\Theta_i + \mathrm{tr}(S_i \Theta_i) + \frac{1}{|P_i|}\|\lambda \circ \Theta_i\|_1 \nonumber\\
&\mathrm{subject\ to}\qquad \Theta_i \in \mathcal{T}.
\label{Toeplitz}
\end{align}
Problem \eqref{Toeplitz} is a convex optimization problem, which we call the \emph{Toeplitz graphical lasso}. This is a variation on the well-known graphical lasso problem~\cite{FHT:08} where we add a block Toeplitz constraint on the inverse covariance.  
The original graphical lasso defines a tradeoff between two objectives, regulated by the parameter $\lambda$: minimizing the negative log likelihood, and making sure $\Theta_i$ is sparse.
When $S_i$ is invertible, the likelihood term encourages $\Theta_i$ to be near $S_i^{-1}$.
Our problem adds the additional constraint that $\Theta_i$ is block Toeplitz.
$\lambda$ is a $nw \times nw$ matrix, so it can be used to regularize each sub-block of $\Theta_i$ differently. 
Note that $\frac{1}{|P_i|}$ can be incorporated into the regularization by simply scaling $\lambda$; as such, we typically write Problem \eqref{Toeplitz} without this term (and scale $\lambda$ accordingly) for notational simplicity.

\section{TICC Algorithm}
\label{sec:proposed}

Here, we describe our algorithm to cluster $X_1, \ldots, X_T$ into $K$ clusters.
Our method, described in full in Section 4.3, depends on two key subroutines: AssignPointsToClusters, where we use a dynamic programming algorithm to assign each $X_t$ into a cluster, and UpdateClusterParameters, where we update the cluster parameters by solving the Toeplitz graphical lasso problem using an algorithm based on the alternating direction method of multipliers (ADMM). 
Note that this is similar to expectation maximization (EM), with the two subroutines corresponding to the E and M steps, respectively.

\subsection{Cluster Assignment}

\begin{figure}[t]
  \centering
  \includegraphics[width=\linewidth]{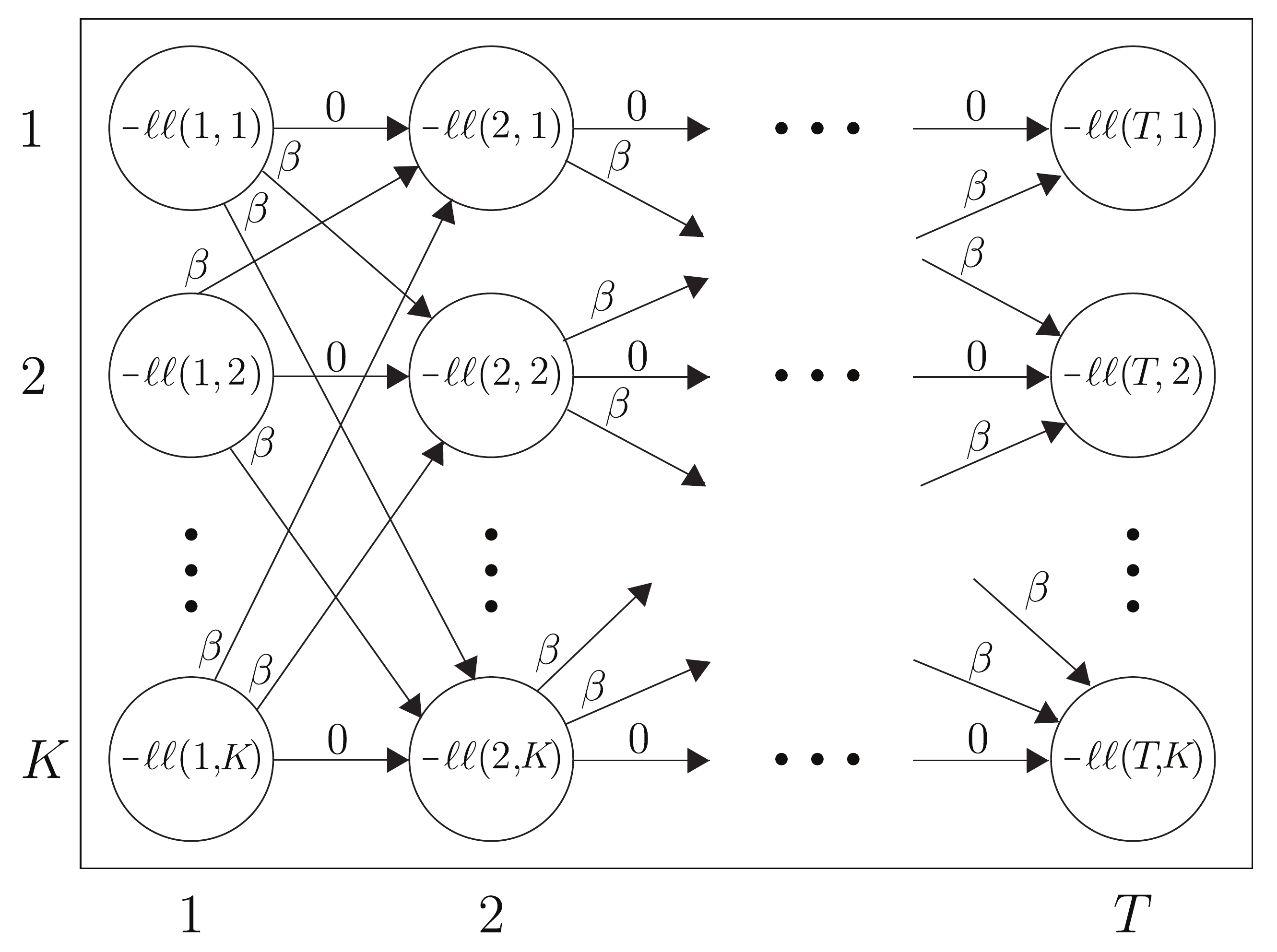}
  \vspace{-8mm}
  \caption{Problem \eqref{clustAssign} is equivalent to finding the minimum-cost path from timestamp $1$ to $T$, where the node cost is the negative log likelihood of that point being assigned to a given cluster, and the edge cost is $\beta$ whenever the cluster assignment switches.}
  \vspace{-2mm}
  \label{fig:viterbi}
\end{figure}

Given the model parameters (\ie, inverse covariances) for each of the $K$ clusters, solving Problem \eqref{clustAssign} assigns the $T$ subsequences, $X_1,\ldots,X_T$, to these $K$ clusters in such a way that maximizes the likelihood of the data while also minimizing the number of times that the cluster assignment changes across the time series. 
Given $K$ potential cluster assignments of the $T$ points, this combinatorial optimization problem has $K^{T}$ possible assignments of points to clusters, that it can choose from. However, we are able to solve for the globally optimal solution in only $O(KT)$ operations.
We do so through a dynamic programming method described in Algorithm \eqref{alg:dynamic}.
This method is equivalent to finding the minimum cost Viterbi path~\cite{viterbi1967error} for this length-$T$ sequence, as visualized in Figure \ref{fig:viterbi}. 

\begin{algorithm}[t]
\caption{Assign Points to Clusters}
\label{alg:dynamic}
\begin{algorithmic}[1]
\small
\State \textbf{given} $\beta > 0$, $-\ell\ell(i,j)$ = negative log likelihood of point $i$ when it is assigned to cluster $j$.
\State \textbf{initialize} PrevCost = list of $K$ zeros. 
\State \qquad \qquad CurrCost = list of $K$ zeros.
\State \qquad \qquad PrevPath = list of $K$ empty lists.
\State \qquad \qquad CurrPath = list of $K$ empty lists.
\For{$i = 1, \ldots,$ $T$}
\For{$j = 1, \ldots, K$}
\State MinIndex = index of minimum value of PrevCost.
\If{PrevCost[MinIndex] + $\beta >$ PrevCost[$j$]}
\State CurrCost[$j$] = PrevCost[$j$] $- \ell\ell(i,j)$.
\State CurrPath[$j$] = PrevPath[$j$].append[$j$].
\Else
\State CurrCost[$j$] = PrevCost[minIndex] + $\beta - \ell\ell(i,j)$.
\State CurrPath[$j$] = PrevPath[minIndex].append[$j$].
\EndIf
\EndFor
\State PrevCost = CurrCost.
\State PrevPath = CurrPath.
\EndFor
\State FinalMinIndex = index of minimum value of CurrCost.
\State FinalPath = CurrPath[FinalMinIndex].
\State \textbf{return} FinalPath.
\end{algorithmic}
\end{algorithm}

\subsection{Solving the Toeplitz Graphical Lasso}

Once we have the clustering assignments, the M-step of our EM algorithm is to update the inverse covariances, given the points assigned to each cluster. Here, we are solving the Toeplitz graphical lasso, which is defined in Problem \eqref{Toeplitz}. 
For smaller covariances, this semidefinite programming problem can be solved using standard interior point methods~\cite{BV:04, OCPB:16}. However, to solve the overall TICC problem, we need to solve a separate Toeplitz graphical lasso for each cluster at every iteration of our algorithm. Therefore, since we may need to solve Problem \eqref{Toeplitz} hundreds of times before TICC converges, it is necessary to develop a fast method for solving it efficiently.
We do so through the alternating direction method of multipliers (ADMM), a distributed convex optimization approach that has been shown to perform well at large-scale optimization tasks~\cite{BPCPE:11, PB:14}. With ADMM, we split the problem up into two subproblems and use a message passing algorithm to iteratively converge to the globally optimal solution. ADMM is especially scalable when closed-form solutions can be found for the ADMM subproblems, which we are able to derive for the Toeplitz graphical lasso. 

To put Problem \eqref{Toeplitz} in ADMM-friendly form, we introduce a consensus variable $Z$ and rewrite Problem \eqref{Toeplitz} as its equivalent problem,
\begin{align*}
\small
&\mathrm{minimize}\qquad -\log\det\Theta + \mathrm{tr}(S \Theta) + \|\lambda \circ Z\|_1 \nonumber\\
&\mathrm{subject\ to}\qquad \Theta = Z, Z \in \mathcal{T}.
\end{align*}

The augmented Lagrangian~\cite{H:69} can then be expressed as
\begin{align}
\small
\mathcal{L}_{\rho}(\Theta,Z,U) := &- \log\det(\Theta) + \Tr{S \Theta} + \|\lambda \circ Z\|_1 \nonumber\\
&+\frac{\rho}{2} \fronorm{\Theta - Z + U}^2.
\label{augLag}
\end{align}
where $\rho > 0$ is the ADMM penalty parameter, $U \in \reals^{nw\times nw}$ is the scaled dual variable ~\cite[\S 3.1.1]{BPCPE:11}, and $Z \in \mathcal{T}$.

ADMM consists of the following three steps repeated until convergence,
\begin{align*}
\small
(a) &\quad \Theta^{k+1} := \argmin{\Theta}\mathcal{L}_\rho\left(\Theta, Z^{k}, U^{k}\right)\\
(b) &\quad Z^{k+1} := \argmin{Z \in \mathcal{T}}\mathcal{L}_\rho\left(\Theta^{k+1}, Z, U^{k}\right)\\
(c) &\quad U^{k+1} := U^k + (\Theta^{k+1} - Z^{k+1}),
\end{align*} 
where $k$ is the iteration number. Here, we alternate optimizing Problem \eqref{augLag} over $\Theta$ and then over $Z$, and after each iteration we update the scaled dual variable $U$. Since the Toeplitz graphical lasso problem is convex, ADMM is guaranteed to converge to the global optimum. We use a stopping criteria based on the primal and dual residual values being close to zero; see~\cite{BPCPE:11}.

\xhdr{$\Theta$-Update}
The $\Theta$-update can be written as
\begin{align*}
\small
\Theta^{k+1} = \argmin{\Theta} - \log\det(\Theta) + \Tr{S \Theta} + \frac{\rho}{2} \fronorm{\Theta - Z^{k} + U^{k}}^2.
\end{align*}

This optimization problem has a known analytical solution~\cite{DWW:14},
\begin{align}
\small
\Theta^{k+1} &= \frac{1}{2\rho}Q\big(D+\sqrt{D^2 + 4\rho I}\big)Q^T,
\end{align}
where $QDQ^T$ is the eigendecomposition of $\rho(Z^{k} - U^{k}) - S$.

\xhdr{Z-Update}
The $Z$-update can be written as
\begin{align}
\small
Z^{k+1} = \argmin{Z \in \mathcal{T}}  \|\lambda \circ Z\|_1 + \frac{\rho}{2} \fronorm{Z - \Theta^{k+1} - U^{k}}^2.
\label{ZproxOp}
\end{align}
This proximal operator can be solved in parallel for each sub-block $A^{(0)}, A^{(1)}, \ldots, A^{(w-1)}$. Furthermore, within each sub-block, each $(i,j)$-th element in the sub-block can be solved in parallel as well. 
In $A^{(0)}$, there are $\frac{n(n+1)}{2}$ independent problems (since it is symmetric), whereas for the other $w-1$ blocks, this number is $n^2$. Therefore, Problem \eqref{ZproxOp} can be broken down into a total of $(w-1)n^2 + \frac{n(n+1)}{2}$ independent problems. Each of these problems has the same form, and can be solved in the same way.

We denote the number of times each element appears as $R$ (this equals 
$2(w-m)$ for sub-block $A^{(m)}$, except for the diagonals of $A^{(0)}$, which occur $w$ times.)
We order these $R$ occurrences, and we let $B^{(m)}_{ij,l}$ refer to the index in $Z$ corresponding to the $l$-th occurrence of the $(i,j)$-th element of $A^{(m)}$, where $l = 1,\ldots, R$. 
Thus, $B^{(m)}_{ij,l}$ returns an index  $(x,y)$ in the $nw \times nw$ matrix of $Z$. 
With this notation, we can solve each of the $(w-1)n^2 + \frac{n(n+1)}{2}$ subproblems of the $Z$-update proximal operator the same way. 
To solve for the elements of $Z$ corresponding to $B^{(m)}_{ij}$, we set these elements all equal to
\begin{align}
\small
\argmin{z} \sum_{l = 1}^{R} |\lambda_{B^{(m)}_{ij,l}}z| + \frac{\rho}{2} \left( z - (\Theta^{k+1} + U^{k})_{B^{(m)}_{ij,l}} \right)^2.
\label{zUpdatePt1}
\end{align}
We let $Q = \sum_{l = 1}^{R} \lambda_{B^{(m)}_{ij,l}}$ and $S_l = (\Theta^{k+1} + U^{k})_{B^{(m)}_{ij,l}}$ for notational simplicity. Then, Problem \eqref{zUpdatePt1} can be rewritten as
\begin{align*}
\small
\argmin{z}  Q |z| + \sum_{l = 1}^{R} \frac{\rho}{2} (z - S_l)^2.
\end{align*}
This is just a soft-threshold proximal operator~\cite{PB:14}, which has the following closed-form solution,
\begin{align}
\small
Z_{B^{(m)}_{ij}}^{k+1} = 
\begin{cases}
\frac{\rho \sum_{l} S_l - Q}{\rho R} \quad & \frac{ \rho \sum_{l} S_l - Q}{\rho R} > 0\\
\frac{\rho \sum_{l} S_l + Q}{\rho R} \quad & \frac{ \rho \sum_{l} S_l + Q}{\rho R} < 0\\
0 \quad & \mbox{otherwise}.
\end{cases}
\end{align}
We fill in the $R$ elements in $Z^{k+1}$ (corresponding to $B^{(m)}_{ij}$) with this value. We do the same for all $(w-1)n^2 + \frac{n(n+1)}{2}$ subproblems, each of which we can solve in parallel, and we are left with our final result for the overall $Z$-update.

\subsection{TICC Clustering Algorithm}

To solve the TICC problem, we combine the dynamic programming algorithm from Section 4.1 and the ADMM method in Section 4.2 into one iterative EM algorithm. We start by randomly initializing the clusters. From there, we alternate the E and M-steps until the cluster assignments are stationary (\ie, the problem has converged). The overall TICC method is outlined in Algorithm \eqref{ticcAlgo}

\begin{algorithm}[t]
\caption{Toeplitz Inverse Covariance-Based Clustering}
\label{ticcAlgo}
\begin{algorithmic}[1]
\small
\State \textbf{initialize} Cluster parameters $\mathbf{\Theta}$; cluster assignments $\mathbf{P}$.
\Repeat 
\State \emph{E-step:} Assign points to clusters \hspace{2.5mm} $\rightarrow \mathbf{P}$. 
\Until{Stationarity.} \newline
\Return $(\mathbf{\Theta}, \mathbf{P}).$ 
\end{algorithmic}
\end{algorithm}

\section{Implementation}
\label{sec:implementation}

We have built a custom Python solver to run the TICC algorithm\footnote{Code and solver are available at \url{http://snap.stanford.edu/ticc/}.}. Our solver takes as inputs the original multivariate time series and the problem parameters. It then returns the clustering assignments of each point in the time series, along with the structural MRF representation of each cluster.

\section{Experiments}
\label{sec:experiments}

We test our TICC method on several synthetic examples. We do so because there are known ``ground truth'' clusters to evaluate the accuracy of our method. 

\xhdr{Generating the Datasets}
We randomly generate synthetic multivariate data in $\reals^5$. Each of the $K$ clusters has a mean of $\vec{0}$ so that the clustering result is based entirely on the structure of the data. For each cluster, we generate a random ground truth Toeplitz inverse covariance as follows~\cite{MLFWL:14}:
\begin{enumerate}[nolistsep, leftmargin=.2in]
    \item Set $A^{(0)}, A^{(1)}, \ldots A^{(4)} \in \reals^{5\times5}$ equal to the adjacency matrices of $5$ independent Erd\H{o}s-R\'enyi directed random graphs, where every edge has a 20\% chance of being selected.
    \item For every selected edge in $A^{(m)}$ set $A^{(m)}_{jk} = v_{jk,m}$, a random weight centered at $0$ (For the $A^{(0)}$ block, we also enforce a symmetry constraint that every $A^{(0)}_{ij} = A^{(0)}_{ji}$). 
    \item Construct a $5w \times 5w$ block Toeplitz matrix $G$, where $w=5$ is the window size, using the blocks $A^{(0)}, A^{(1)}, \ldots A^{(4)}$.
    \item Let $c$ be the smallest eigenvalue of $G$, and set $\Theta_i = G + (0.1+|c|)I$. This diagonal term ensures that $\Theta_i$ is invertible.
\end{enumerate}
The overall time series is then generated by constructing a temporal sequence of cluster segments (for example, the sequence ``$1,2,1$'' with $200$ samples in each of the three segments, coming from two inverse covariances $\Theta_1$ and $\Theta_2$). The data is then drawn one sample at a time, conditioned on the values of the previous $w-1$ samples. Note that, when we have just switched to a new cluster, we are drawing a new sample in part based on data that was generated by the previous cluster.

We run our experiments on four different temporal sequences: ``1,2,1'', ``1,2,3,2,1'', ``1,2,3,4,1,2,3,4'', ``1,2,2,1,3,3,3,1''. Each segment in each of the examples has $100K$ observations in $\reals^5$, where $K$ is the number of clusters in that experiment (2, 3, 4, and 3, respectively). These examples were selected to convey various types of temporal sequences over various lengths of time.

\xhdr{Performance Metrics}
We evaluate performance by clustering each point in the time series and comparing to the ground truth clusters. 
Since both TICC and the baseline approaches use very similar methods for selecting the appropriate number of clusters, we fix $K$ to be the ``true'' number of clusters, for both TICC and for all the baselines.
This yields a straightforward multiclass classification problem, which allows us to evaluate clustering accuracy by measuring the macro-$F_1$ score. For each cluster, the $F_1$ score is the harmonic mean of the precision and recall of our estimate. Then, the macro-$F_1$ score is the average of the $F_1$ scores for all the clusters. We use this score to compare our TICC method with several well-known time series clustering baselines.

\xhdr{Baseline Methods}
We use multiple model and distance-based clustering approaches as our baselines. The methods we use are:
\begin{itemize}[nolistsep, leftmargin=.2in]
    \item TICC, $\beta=0$ --- This is our TICC method without the temporal consistency constraint. Here, each subsequence is assigned to a cluster independently of its location in the time series. 
    \item GMM --- Clustering using a Gaussian Mixture Model~\cite{banfield1993model}.
    \item EEV --- Regularized GMM with shape and volume constraints on the Gaussian covariance matrix~\cite{fraley2006mclust}.
    \item DTW, GAK --- Dynamic time warping (DTW)-based clustering using a global alignment kernel~\cite{cuturi2011fast,dtwclust}.
    \item DTW, Euclidean --- DTW using a Euclidean distance metric~\cite{dtwclust}.
    \item Neural Gas --- Artificial neural network clustering method, based on self-organizing maps~\cite{dimtriadou2009cclust,martinetz1993neural}.
    \item K-means --- The standard K-means clustering algorithm using Euclidean distance.
\end{itemize}

\begin{table}[t]
\centering
\resizebox{8.6cm}{!} {
    \begin{tabular}{c | c || c | c | c | c }
    & & \multicolumn{4}{c}{Temporal Sequence} \\ 
    \hline
    & Clustering Method & 1,2,1 & 1,2,3,2,1 & 1,2,3,4,1,2,3,4 & 1,2,2,1,3,3,3,1 \\
    \hhline{=|=||=|=|=|=}
    & \textbf{TICC} & \textbf{0.92} & \textbf{0.90} & \textbf{0.98} & \textbf{0.98} \\
    \hline
    \multirow{3}{*}{Model-} & TICC, $\beta=0$ & 0.88 & 0.89 & 0.86 & 0.89 \\\cline{2-2}
    \multirow{3}{*}{Based} & GMM & 0.68 & 0.55 & 0.83 & 0.62 \\ \cline{2-2}
    \multirow{3}{*}{} & EEV & 0.59 & 0.66 & 0.37 & 0.88 \\ 
    \hline     
    \multirow{4}{*}{Distance-} & DTW, GAK & 0.64 & 0.33 & 0.26 & 0.27 \\ \cline{2-2}
    \multirow{4}{*}{Based} & DTW, Euclidean & 0.50 & 0.24 & 0.17 & 0.25 \\ \cline{2-2}
    \multirow{4}{*}{} & Neural Gas & 0.52 & 0.35 & 0.27 & 0.34\\ \cline{2-2}
    \multirow{4}{*}{} & K-means & 0.59 & 0.34 & 0.24 & 0.34 \\
    \hline

    \end{tabular}
    }
    \caption{Macro-$F_1$ score of clustering accuracy for four different temporal sequences, comparing TICC with several alternative model and distance-based methods.}
    \label{synTable}
    \vspace{-8mm}
\end{table}

\begin{figure}[t]
  \centering
  \includegraphics[width=\linewidth]{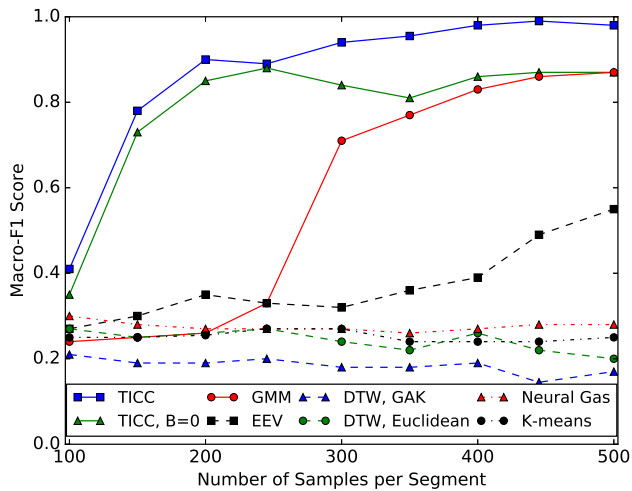}
  \vspace{-6mm}
  \caption{Plot of clustering accuracy macro-$F_1$ score vs. number of samples for TICC and several baselines. TICC needs significantly fewer samples than the other model-based methods to achieve similar performance, while the distance-based measures are unable to capture the true structure.}
  \label{fig:robustness}
  \vspace{-6mm}
\end{figure}

\xhdr{Clustering Accuracy}
We measure the macro-$F_1$ score for the four different temporal sequences in Table \ref{synTable}. Here, all eight methods are using the exact same synthetic data, to isolate each approach's effect on performance. As shown, TICC significantly outperforms the baselines. Our method achieves a macro-$F_1$ score between $0.90$ and $0.98$, averaging $0.95$ across the four examples. This is $41\%$ higher than the second best method (not counting TICC, $\beta=0$), which is GMM and has an average macro-$F_1$ score of only $0.67$. We also ran our experiments using micro-$F_1$ score, which uses a weighted average to weigh clusters with more samples more heavily, and we obtained very similar results (within 1-2\% of the macro-$F_1$ scores).
Note that the $K$ clusters in our examples are always zero-mean, and that they are only differentiated by the structure of the data. As a result, the distance-based methods struggle at identifying the clusters, and these approaches have lower scores than the model-based methods for these experiments.

\xhdr{Effect of the Total Number of Samples}
We next focus on how many samples are required for each method to accurately cluster the time series. We take the ``1,2,3,4,1,2,3,4'' example from Table \ref{synTable} and vary the number of samples.
We plot the macro-$F_1$ score vs. number of samples per segment for each of the eight methods in Figure \ref{fig:robustness}.
As shown, when there are 100 samples, none of the methods are able to accurately cluster the data. However, as we observe more samples, both TICC and TICC, $\beta=0$ improve rapidly. By the time there are 200 samples, TICC already has a macro-$F_1$ score above 0.9. 
Even when there is a limited amount of data, our TICC method is still able to accurately cluster the data. Additionally, we note that the temporal consistency constraint, defined by $\beta$, has only a small effect in this region, since both TICC and TICC, $\beta=0$ achieve similar results. 
Therefore, the accurate results are most likely due to the sparse block Toeplitz constraint that we impose in our TICC method.
However, as the number of samples increases, these two plots begin to diverge, as TICC goes to 1.0 while TICC, $\beta=0$ hovers around 0.9.
This implies that, once we have enough samples, the final improvement in performance is due to the temporal consistency penalty that encourages neighboring samples to be assigned to the same cluster.

\xhdr{Network Recovery Accuracy}
Recall that our TICC method has the added benefit in that the clusters it learns are interpretable. TICC models each cluster as a multilayer Markov random field, a network with edges corresponding to the non-zero entries in the inverse covariance matrix $\Theta_i$. We can compare our estimated network with the ``true'' MRF network and measure the average macro-$F_1$ score of our estimate across all the clusters. 
We look at the same four examples as in Table \ref{fig:network} and plot the results in Table \ref{fig:network}.
We recover the underlying edge structure of the network with an $F_1$ score between 0.79 and 0.90. 
This shows that TICC is able to both accurately cluster the data and recover the network structure of the underlying clusters.
Note that our method is the first approach that is able to explicitly reconstruct this network, something that the other baseline methods are unable to do.

\begin{table}[t]
\centering
    \begin{tabular}{c | c }
	Temporal Sequence & TICC Network Recovery $F_1$ score \\
    \hline
 	 1,2,1 & 0.83 \\
 	 1,2,3,2,1 & 0.79 \\
  	 1,2,3,4,1,2,3,4 & 0.89 \\
  	 1,2,2,1,3,3,3,1 & 0.90 \\
    \hline
    \end{tabular}
    \caption{Network edge recovery $F_1$ score for the four temporal sequences. TICC defines each cluster as an MRF graphical model, which is successfully able to estimate the dependency structure of the underlying data.}
    \label{fig:network}
    \vspace{-8mm}
\end{table}

\xhdr{Window Size Robustness}
We next examine how the selection of window size $w$ affects our results. We run the same ``1,2,3,4,1,2,3,4'' example, except now we vary the window size $w$. (Recall that the ``true'' window size was 5.) Empirically, we discover that any window size between 4 and 15 yields a Macro-$F_1$ clustering accuracy score of between 0.95 and 0.98. Similarly, our network recovery macro-$F_1$ score stays between 0.87 and 0.89 for window sizes between 5 and 14. It is only after the window size drops below 4 or above 15 that the results begin to get worse. We observe similar patterns in the other three examples, so our TICC method appears to be relatively robust to the selection of $w$.

\begin{figure}[t]
  \centering
  \includegraphics[width=\linewidth]{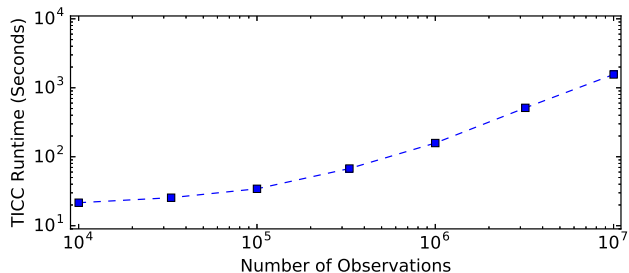}
  \vspace{-8mm}
  \caption{Per-iteration runtime of the TICC algorithm (both the ADMM and dynamic programming steps). Our algorithm scales linearly with the number of samples. In this case, each observation is a vector in $\reals^{50}$.}
  \label{fig:scalability}
  \vspace{-4mm}
\end{figure}

\xhdr{Scalability of TICC}
One iteration of the TICC algorithm consists of running the dynamic programming algorithm and then solving the Toeplitz graphical lasso problem for each cluster. These steps are repeated until convergence. The total number of iterations depends on the data, but typically is no more than a few tens of iterations.
Since $T$ is typically much larger than both $K$ and $n$, we can expect the largest bottleneck to occur during the assignment phase, where $T$ can potentially be in the millions.
To evaluate the scalability of our algorithm, we vary $T$ and compute the runtime of the algorithm over one iteration. We observe samples in $\reals^{50}$, estimate 5 clusters with a window size of 3, and vary $T$ over several orders of magnitude. We plot the results in log-log scale in Figure \ref{fig:scalability}. Note that our ADMM solver infers each $150\times150$ inverse covariance (since $nw = 50\times3=150$) in under 4 seconds, but this runtime is independent of $T$, so ADMM contributes to the constant offset in the plot. As shown, at large values of $T$, our algorithm scales linearly with the number of points. Our TICC solver can cluster 10 millions points, each in $\reals^{50}$, with a per-iteration runtime of approximately 25 minutes.

\section{Case Study}
\label{sec:applications}

Here, we apply our TICC method to a real-world example to demonstrate how this approach can be used to find meaningful insights from time series data in an unsupervised way.

We analyze a dataset, provided by a large automobile company, containing sensor data from a real driving session. This session lasts for exactly 1 hour and occurs on real roads in the suburbs of a large European city. We observe 7 sensors every 0.1 seconds:
    \vspace{-1\baselineskip}
	\setlength{\columnsep}{-0.3cm}
    \begin{multicols}{2}
    \begin{itemize}[nolistsep]
        \item Brake Pedal Position
        \item Forward (X-)Acceleration
        \item Lateral (Y-)Acceleration
        \item Steering Wheel Angle
        \item Vehicle Velocity
        \item Engine RPM
        \item Gas Pedal Position
    \end{itemize}
    \end{multicols}
	\setlength{\columnsep}{1cm}
    \vspace{-1\baselineskip}
Thus, in this one-hour session, we have 36,000 observations of a 7-dimensional time series. We apply TICC with a window size of 1 second (or 10 samples). We pick the number of clusters using BIC, and we discover that this score is optimized at $K = 5$.

We analyze the 5 clusters outputted by TICC to understand and interpret what ``driving state'' they each refer to. Each cluster has a multilayer MRF network defining its structure. To analyze the result, we use network analytics to determine the relative ``importance'' of each node in the cluster's network. We plot the betweenness centrality score~\cite{brandes2001faster} of each node in Table \ref{betweenness}. We see that each of the 5 clusters has a unique ``signature'', and that different sensors have different betweenness scores in each cluster. For example, the Y-Acceleration sensor has a non-zero score in only two of the five clusters: \#2 and \#5. As such, we would expect these two clusters to refer to states in which the car is turning, and the other three clusters to refer to intervals where the car is going straight. Similarly, cluster \#1 is the only cluster with no importance on the Gas Pedal, and it is also the cluster with the largest Brake Pedal score. Therefore, we expect this state to be the cluster assignment whenever the car is slowing down. We also see that clusters 3 and 4 have the same non-zero sensors, velocity and gas pedal, so we may expect them to refer to states when the car is driving straight and not slowing down, with the most ``important'' sensor in the two clusters being the velocity in cluster 4. As such, we can use these betweenness scores to interpret these clusters in a meaningful way.
For example, from Table \ref{betweenness}, a reasonable hypothesis might be that the clusters refer to 1) slowing down, 2) turning, 3) speeding up, 4) cruising on a straight road, 5) driving on a curvy road segment.

\begin{table}[t]
\centering
\resizebox{8.6cm}{!} {
    \begin{tabular}{c | c || c | c | c | c | c | c | c }
	 & Interpretation & Brake & X-Acc & Y-Acc & SW Angle & Vel & RPM & Gas \\
    \hhline{=|=||=|=|=|=|=|=|=}
    \#1 & Slowing Down & 25.64 & 0 & 0 & 0 & 27.16 & 0 & 0 \\
    \hline
    \#2 & Turning & 0 & 4.24 & 66.01 & 17.56 & 0 & 5.13 & 135.1 \\
    \hline
    \#3 & Speeding Up & 0 & 0 & 0 & 0 & 16.00 & 0 & 4.50 \\
    \hline
    \#4 & Driving Straight & 0 & 0 & 0 & 0 & 32.2 & 0 & 26.8 \\
    \hline
    \#5 & Curvy Road & 4.52 & 0 & 4.81 & 0 & 0 & 0 & 94.8 \\
    \hline
    \end{tabular}
    }
    \caption{Betweenness centrality for each sensor in each of the five clusters. This score can be used as a proxy to show how ``important'' each sensor is, and more specifically how much it directly affects the other sensor values.}
    \label{betweenness}
   \vspace{-10mm}
\end{table}

\begin{figure}[!t]
\centering 
  \subfigure[]{\label{fig:turn1}\centering\includegraphics[width=0.525\linewidth]{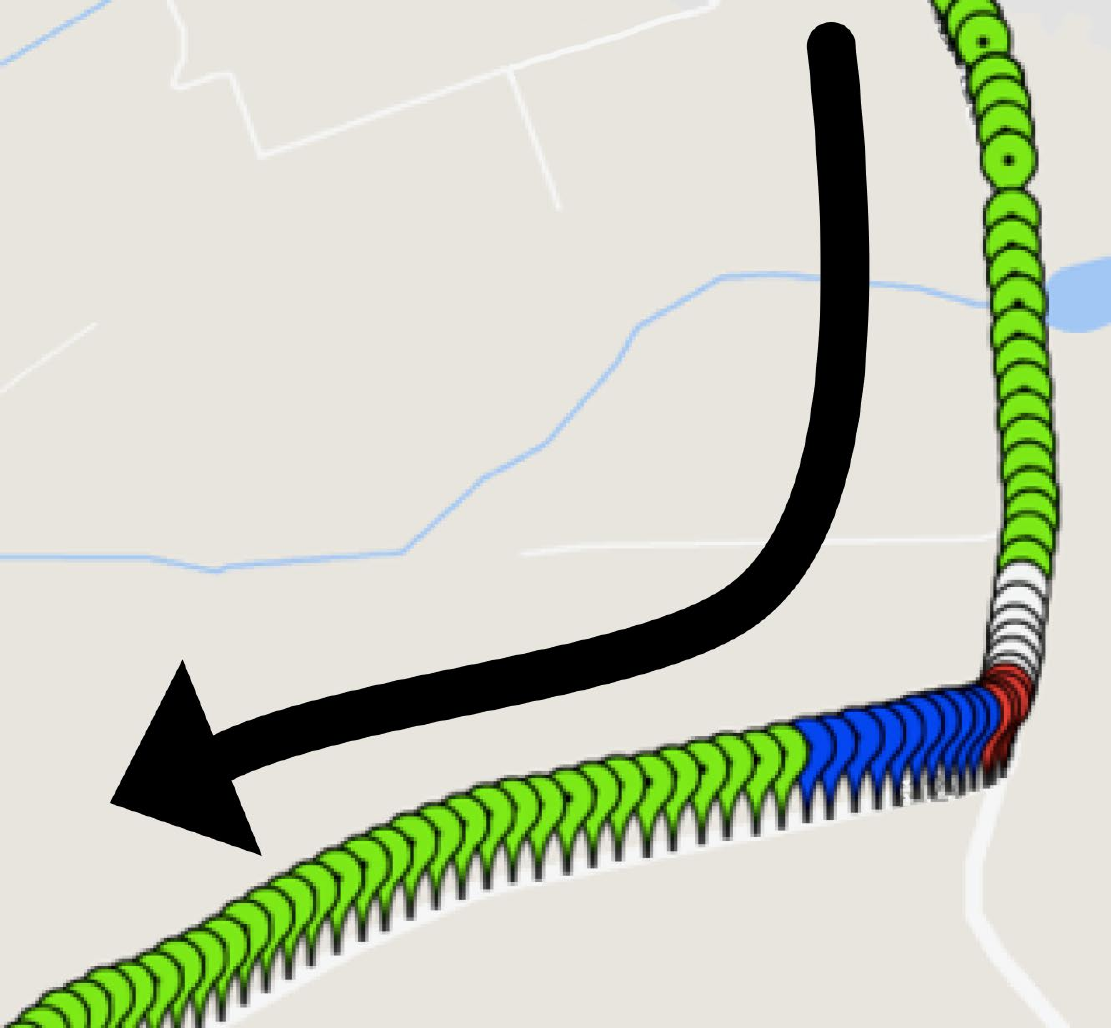}}
  \subfigure[]{\label{fig:turn2}\centering\includegraphics[width=0.46\linewidth]{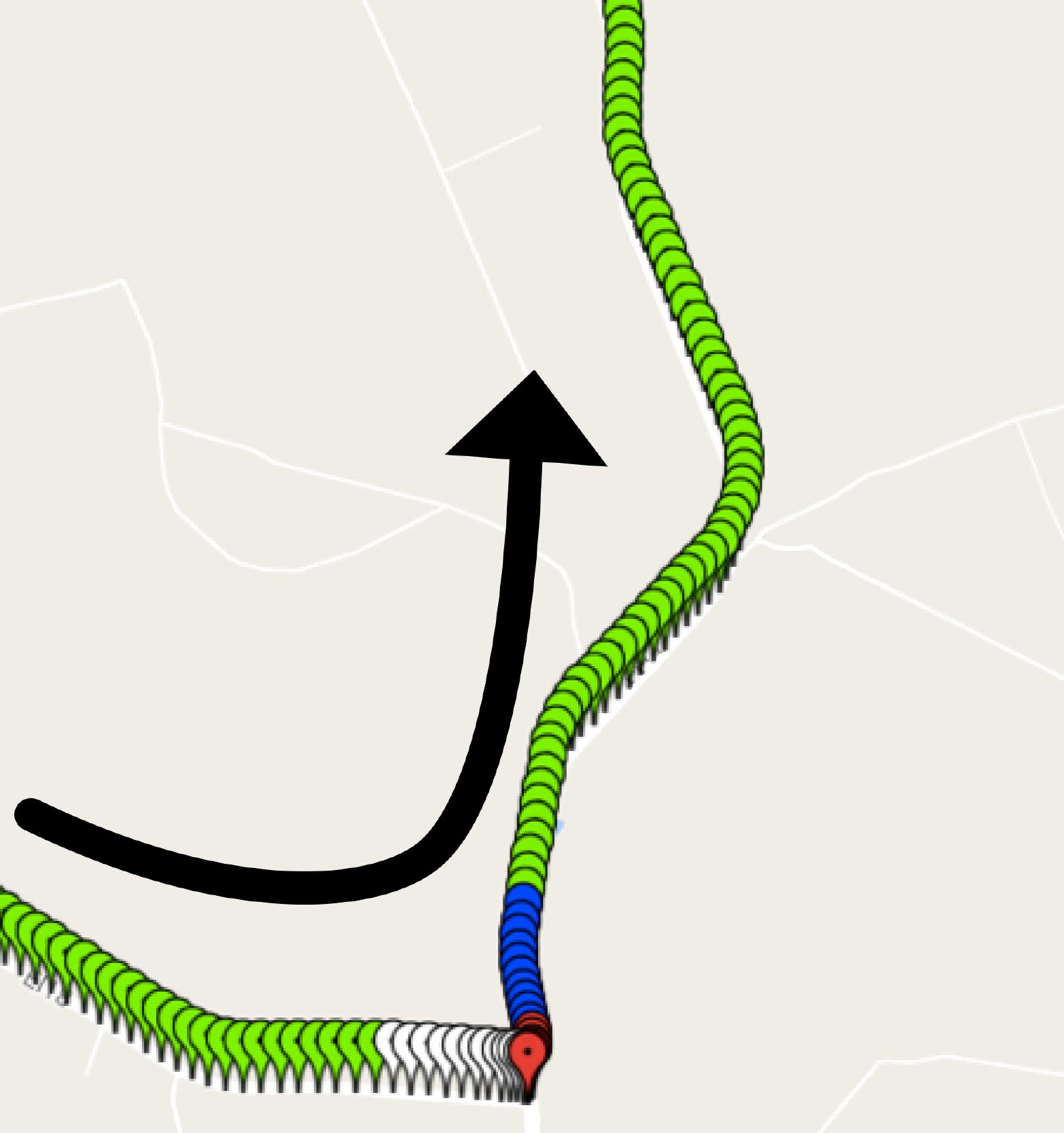}}
  \vspace{-4mm}
    \caption{Two real-world turns in the driving session. The pin color represents cluster assignment from our TICC algorithm (Green = Going Straight, White = Slowing Down, Red = Turning, Blue = Speeding up). Since we cluster based on structure, rather than distance, both a left and a right turn look very similar under the TICC clustering scheme.}
   \label{fig:turns}
   \vspace{-6mm}
\end{figure}

\xhdr{Plotting the Resulting Clusters}
To validate our hypotheses, we can plot the latitude/longitude locations of the drive, along with the resulting cluster assignments. Analyzing this data, we empirically discover that each of the five clusters has a clear real-world interpretation that aligns very closely with our estimates based on the betweenness scores in Table \ref{betweenness}.  
Furthermore, we notice that many consistent and repeated patterns emerge in this one hour session. For example, whenever the driver is approaching a turn, he or she follows the same sequence of clusters: going straight, slowing down, turning, speeding up, then going straight again. We plot two typical turns in the dataset, coloring the timestamps according to their cluster assignments, in Figure \ref{fig:turns}. 
It is important to note here that the same pattern emerges here for both left and right turns. Whereas distance-based approaches would treat these two scenarios very differently (since several of the sensors have completely opposite values), TICC instead clusters the time series based on structural similarities. As a result, TICC assigns both left and right turns into the same underlying cluster.

\section{Conclusion and Future Work}
\label{sec:conclusion}

In this paper, we have defined a method of clustering multivariate time series subsequences. Our method, Toeplitz Inverse Covariance-based Clustering (TICC), is a new type of model-based clustering that is able to find accurate and interpretable structure in the data. Our TICC algorithm simultaneously segments and clusters the data, breaking down high-dimensional time series into a clear sequential timeline. We cluster each subsequence based on its correlation structure and define each cluster by a multilayer MRF, making our results highly interpretable. To discover these clusters, TICC alternates between assigning points to clusters in a temporally consistent way, which it accomplishes through dynamic programming, and updating the cluster MRFs, which it does via ADMM. TICC's promising results on both synthetic and real-world data lead to many potential directions for future research. For example, our method could be extended to learn dependency networks parameterized by \emph{any} heterogeneous exponential family MRF. This would allow for a much broader class of datasets (such as boolean or categorical readings) to be incorporated into the existing TICC framework, opening up this work to new potential applications. 

\xhdr{Acknowledgements} This work was supported by NSF IIS-1149837, NIH BD2K, DARPA SIMPLEX, DARPA XDATA, Chan Zuckerberg Biohub, SDSI, Boeing, Bosch, and Volkswagen.

\bibliography{refs}
\bibliographystyle{abbrv}

\end{document}